\documentclass[letterpaper, 10 pt, conference]{ieeeconf}
\IEEEoverridecommandlockouts 
\usepackage{cite}
\usepackage{amsmath,amssymb,amsfonts}
\usepackage{algorithmic}
\usepackage{graphicx}
\usepackage{textcomp}
\usepackage[table,xcdraw]{xcolor}
\usepackage{subcaption}
\usepackage{authblk}
\let\labelindent\relax
\usepackage{enumitem}
\usepackage{hyperref}
\usepackage{breqn}

\newcommand{\nameCAR}{$C$}
\newcommand{\nameLIDARINPUT}{\textbf{S}$_{t}$}
\newcommand{\nameRANGEIMAGE}{\textbf{R}$_{t}$}
\newcommand{\nameCNN}{$F_{\mathrm{cnn}}$}
\newcommand{\nameDRIFT}{$d_t$}
\newcommand{\nameEMBEDDING}{$z_t$}
\newcommand{\nameGCACTIVATION}{\textbf{R}$^{'}_{t}$}

\newcommand{\eqnDTL}{
\begin{equation} \label{eqn:DTL}
    \small
    \mathcal{L} = max(\beta + z_n - z_p + (z_a-z_n)^2 + (z_a-z_p)^2, 0)
\end{equation}
}

\newcommand{\eqnRANGEEXTRACTION}{
\begin{equation} \label{eqn:RIExtraction}
    \small
    r = \sqrt{p_x^2 + p_y^2 + p_z^2}, 
    \phi = \arctan(p_x/p_y), 
    \theta = \arcsin(p_z/r) 
\end{equation}
}

\newcommand{\eqnRANGEPROJECTION}{
\begin{equation} \label{eqn:RIProjection}
    \small
    x_{\mathrm{proj}} = \frac{1}{2} \times \big(\phi/\pi + 1\big),  
    y_{\mathrm{proj}} = 1 - \bigg(\frac{\theta + |\mathrm{fov}_{\mathrm{down}}|}{\mathrm{fov}_{\mathrm{total}}} \bigg), 
\end{equation}
}

\newcommand{\eqnObjectiveFunc}{
\begin{subequations}
\begin{align}
    \small
    \setlength{\abovedisplayskip}{-10pt}
    \min \sum_t c(y_{e, t}, \dot{x}_{e, t}, \ddot{x}_{e,t}, \dot{y}_{e, t}, \ddot{y}_{e, t}, y_{\mathrm{feat}})
    \label{cost_standard} \\
    (x_{e, t_0}, y_{e, t_0}, \dot{x}_{e, t_0}, \dot{y}_{e, t_0}, \ddot{x}_{e, t_0}, \ddot{y}_{e, t_0} ) &= \textbf{b}_0. \label{boundary_cond_initial}\\
    (\dot{x}_{e, t_f}, \dot{y}_{e, t_f}, \ddot{x}_{e, t_f}, \ddot{y}_{e, t_f} ) &= \textbf{b}_f. \label{boundary_cond_final}\\
     f_v(\dot{x}_{e, t}, \dot{y}_{e, t})\leq v_{max}, f_a(\ddot{x}_{e, t}, \ddot{y}_{e, t}) &\leq a_{max} \label{motion_bounds}\\
    -\frac{(x_{e, t}-x_{o, t}^j)^2}{a^2}-\frac{(y_{e, t}-y_{o, t}^j)^2}{b^2}+1 &\leq 0, \label{coll_nonhol_standar}
    \setlength{\belowdisplayskip}{-10pt}
\end{align}
\end{subequations}
\begin{dmath}
    \small
    c(y_{e, t}, \dot{x}_{e, t}, \ddot{x}_{e,t}, \dot{y}_{e, t}, \ddot{y}_{e, t}, y_{\mathrm{feat}}) = (\ddot{x}_{e, t}^2+\ddot{y}_{e, t}^2)+(y_{e, t}-y_{\mathrm{feat}})^2+(\sqrt{\dot{x}_{e, t}^2+\dot{y}_{e, t}^2}-v_{\mathrm{des}})^2\label{cost_function}
\end{dmath}
\vspace*{-0.5mm}
\begin{align}
    \small
    f_v = \sqrt{\dot{x}_{e, t}^2+\dot{y}_{e, t}^2}, f_a = \sqrt{\ddot{x}_{e, t}^2+\ddot{y}_{e, t}^2}
    \label{acc_vel_const_functions}
\end{align}
}

\newcommand{\eqnMetaCost}{
\begin{equation}
    \small
    \setlength{\abovedisplayskip}{2pt}
    c+\max(0, \Vert f_{\mathrm{curv}}\Vert-\kappa_{\mathrm{max}})+\max(0, \Vert \overline{y}_{e, t}^{l}(t)\Vert-y_d),
    \setlength{\belowdisplayskip}{2pt}
    \label{meta_cost}
\end{equation}
}

\newcommand{\tableAblationResults}{
\begin{table}[!htb]
\centering
\caption{\small Ablation Experiment on Feature Density vs. Drift}
\label{tab:AblationResults}
\resizebox{0.8\linewidth}{!}{%
\begin{tabular}{|c|c|c|c|}
\hline
\textbf{Feature Density} & \multicolumn{1}{r|}{\textbf{VC (m)}} & \textbf{LADFN (m)} & \textbf{Ours (m)} \\ \hline
0.25 & 0.818 & 1.12 & \textbf{0.66} \\ \hline
0.16 & 1.05 & 1.27 & \textbf{0.84} \\ \hline
0.125 & 1.45 & 1.50 & \textbf{1.05} \\ \hline
0.1 & 1.14 & 1.23 & \textbf{1.03} \\ \hline
0.083 & 1.344 & \textbf{1.309} & \textbf{1.309} \\ \hline
\end{tabular}%
}
\end{table}
\vspace{-5mm}
}

\newcommand{\captionTeaser}{\small \textbf{Actively moving towards salient semantic features reduces drift.} An ego vehicle in the CARLA \cite{Dosovitskiy17} scene (in \textbf{2}) observes trees on the right and better-defined poles and platforms on the left. \textbf{1} shows the corresponding range image (Section~\ref{sec:RangeImage}) of this scene. \textbf{3} and \textbf{4} show the trajectories executed by a prior approach LADFN \cite{omama2021learning} and our method respectively. By moving towards semantically-better features on the left in \textbf{4}, our approach outperforms LADFN by $\boldsymbol{54.3\%}$, increasing the control cost by only $\boldsymbol{0.4\%}$. Therefore, it is imperative that we identify semantically-better features to minimize drift in active navigation.
}

\newcommand{\captionPipeline}{\small \textbf{Pipeline of our system.} \textbf{1} shows the CARLA \cite{Dosovitskiy17} scene with trees and electric poles on the right side of the ego vehicle \nameCAR. Then, we project a 3D point cloud \nameLIDARINPUT~$\in \mathbb{R}^{3\times N_{\mathrm{pc}}}$~(\textbf{2}) to a 2D range image \nameRANGEIMAGE~$\in \mathbb{R}^{w \times h}$ in \textbf{3}. Then, our neural network perception module \nameCNN~(\textbf{4}) learns to rank these range images based on the drift \nameDRIFT~they accumulate. \textbf{5} shows the GradCAM \cite{selvaraju2017grad} activation \nameGCACTIVATION~$\in \mathbb{R}^{w \times h}$~of the range image shown in \textbf{3}, which highlights the features (seen in dark blue) that are conducive to minimizing drift \nameDRIFT. These features are used by the CEM-MPC Controller \textbf{6} which generates and selects a best-performing trajectory for the ego vehicle \nameCAR~to execute (\textbf{7}). 
}

\newcommand{\captionCNNIpOp}{\small \textbf{Drift minimizing feature detection using GradCAM}. We show a CARLA \cite{Dosovitskiy17} scene in row 1. Its corresponding range image \nameRANGEIMAGE~ is in row 2 and its  feature activation map \nameGCACTIVATION~queried using GradCAM \cite{selvaraju2017grad} in row 3. Pixels in bright blue highlight the drift-minimizing features. \nameGCACTIVATION~is activated more for salient features like poles and tree trunks than planar features like walls. 
}

\newcommand{\captionQualResults}{\small 
\textbf{Our deep-learned model differentiates between drift-minimizing features of several semantic classes.} Each row shows a different CARLA \cite{Dosovitskiy17} scene. Columns 2,3, and 4 show the trajectories executed by VC, LADFN \cite{omama2021learning}, and our approach respectively. Black and red lines show the ground truth trajectories and LOAM's \cite{zhang2014loam} odometry respectively. The zoomed-in circles on top highlight drift. In scenes 1 and 2, we match LADFN \cite{omama2021learning}, both giving an improvement of about $\boldsymbol{51\%}$ over VC in scene 1 and $\boldsymbol{2.9\%}$ in scene 2. In scene 3, our approach outperforms LADFN by $\boldsymbol{54.2\%}$ and VC by $\boldsymbol{42.6\%}$ respectively, because we identify semantically-better features like poles, platforms as opposed to LADFN's \cite{omama2021learning} focus on feature geometry.%
}

\newcommand{\captionQualResultsControl}{\small 
\textbf{Intelligent drift-aware behaviours (left) and their corresponding velocity profiles (right).} 
In green, we show that a yellow ego vehicle will speed up to overtake moving traffic while maintaining proximity to drift-minimizing features on the left. In blue, we show that \textit{without} moving traffic, the ego will smoothly move to the left to achieve a desired velocity of $3 ms^{-1}$. 
}

\newcommand{\captionQuanResults}{\small \textbf{In static scenes}~(rows~1~\&~2), we reduce drift by $\boldsymbol{76.76\%}$ and $\boldsymbol{47.54\%}$ compared to LADFN and VC respectively. Yet, we only increase control cost by $\boldsymbol{0.4\%}$ and $\boldsymbol{2.23\%}$ respectively. \textbf{In dynamic scenes} rows (3 \& 4), we reduce drift by $\boldsymbol{64.22\%}$ and $\boldsymbol{65.17\%}$ compared to the same respective controllers. We \textit{decrease} control cost by $\boldsymbol{0.6\%}$ and increase by $\boldsymbol{1.98\%}$ respectively.%
}

\newcommand{\captionQuanResultsGrad}{\small \textbf{GradCAM (\nameGCACTIVATION) Activation Comparison.} From Cases 2 \& 3 in \ref{sec:QualResults}, \nameGCACTIVATION~is activated more based on which side of the road is populated with semantically-better features like poles, platforms.
}

\newcommand{\SubFigureScalar}{1\textwidth}

\newcommand{\figTeaser}{
\begin{figure}[!htb]
    \includegraphics[width=0.5\textwidth, scale=0.001]{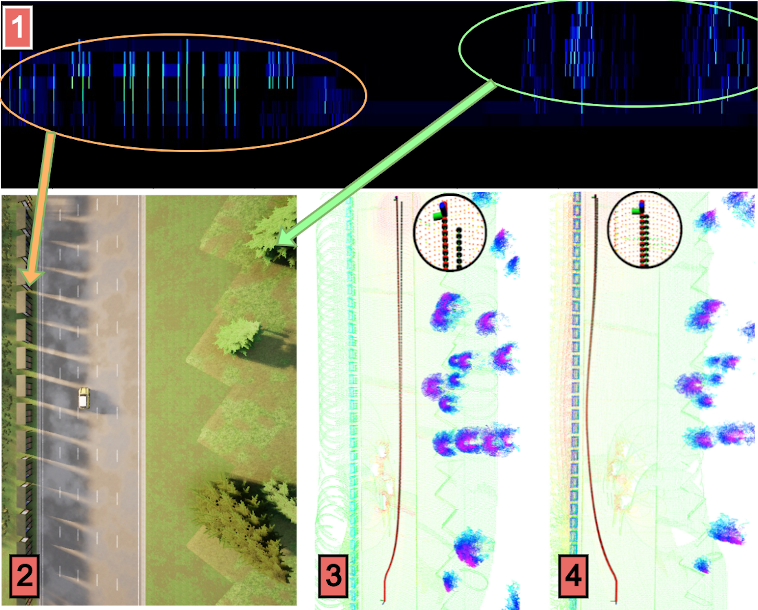}
    \centering
    \setlength{\belowcaptionskip}{-20pt}
    \vspace{-1.8em}
    \caption{\captionTeaser}
    \label{fig:Teaser}
\end{figure}
}

\newcommand{\figPipeline}{
\begin{figure*}[!htb]
    \includegraphics[width=\textwidth]{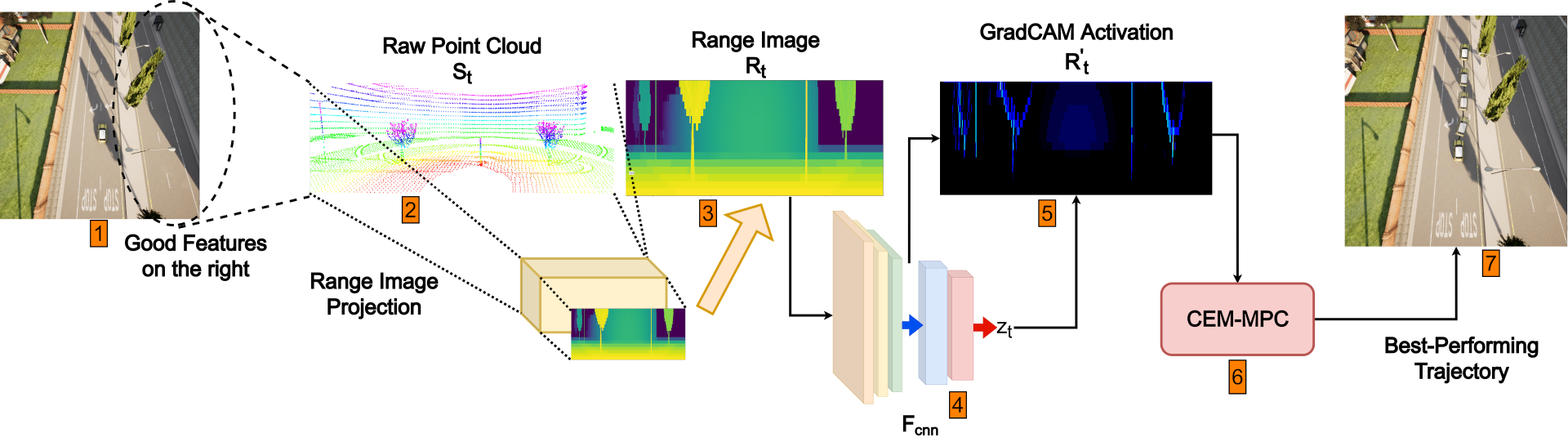}
    \centering
    \setlength{\belowcaptionskip}{-15pt}
    \vspace{-1.7em}
    \caption{\captionPipeline}
    \label{fig:Pipeline}
\end{figure*}
}

\newcommand{\figCNNIpOp}{
\begin{figure}[!htb]
\includegraphics[width=\linewidth, height=0.22\textheight]{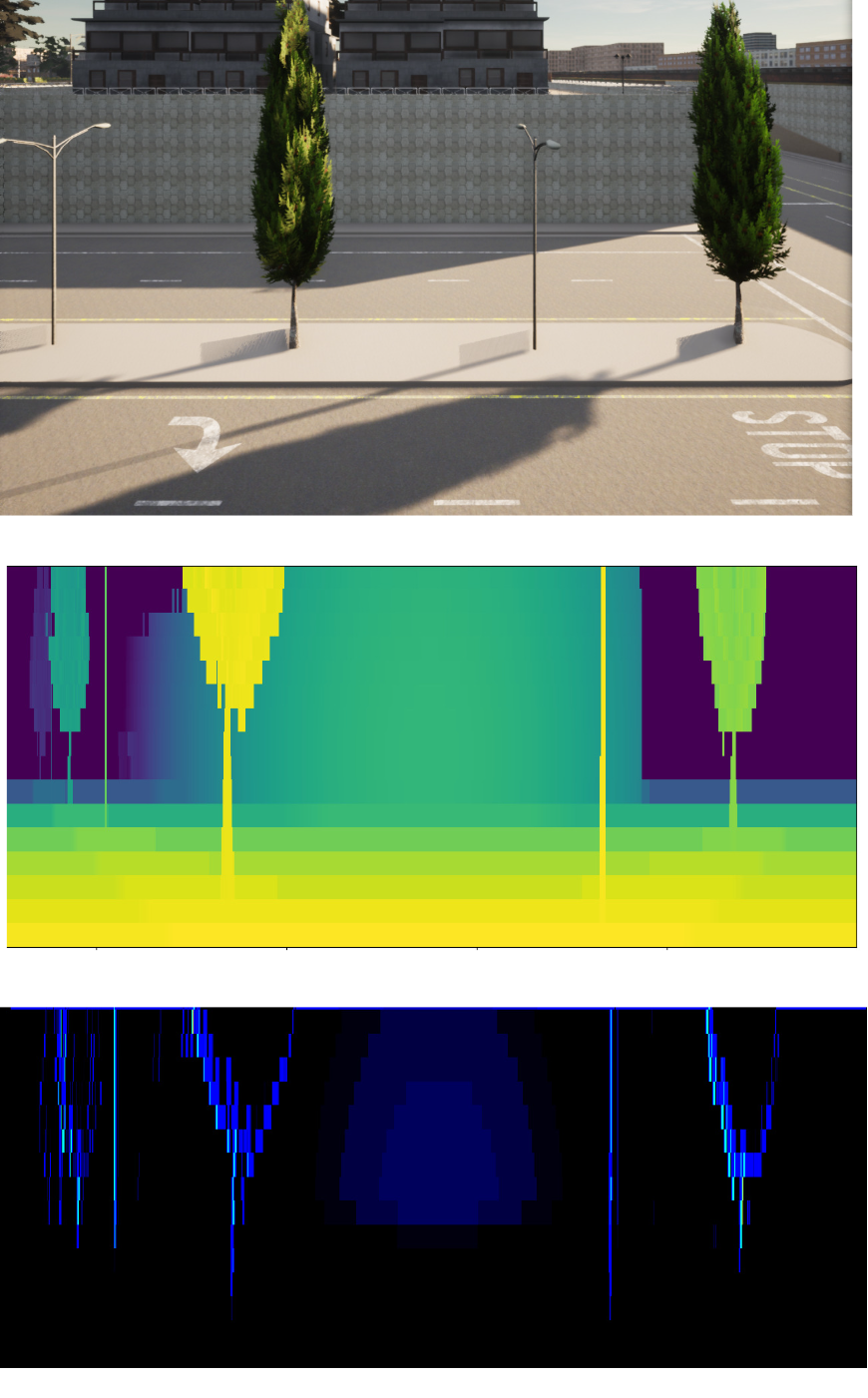}  
\centering
\setlength{\abovecaptionskip}{-10pt}
\setlength{\belowcaptionskip}{-18pt}
\caption{\captionCNNIpOp}
\label{fig:IpOpCNN}
\end{figure}
}

\newcommand{\figCEMPipeline}{
\begin{figure}[!h]
\centering
\includegraphics[width=\linewidth]{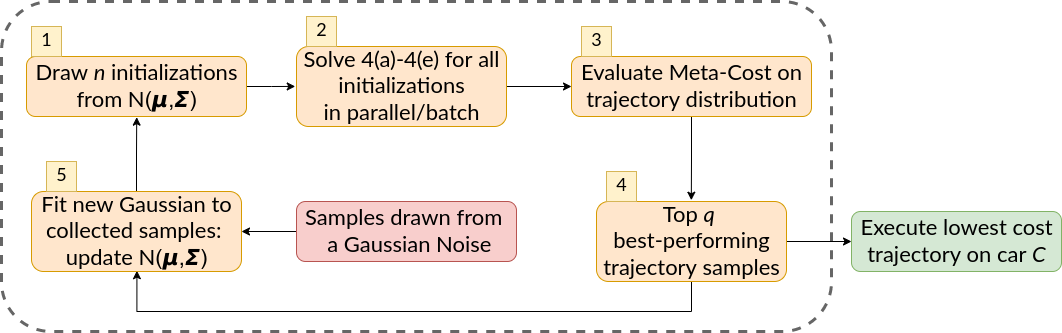}
\setlength{\abovecaptionskip}{-4pt}
\caption{\small \textbf{Our CEM-MPC Pipeline} combines the exploration of CEM with a custom batch non-convex trajectory optimizer.}
\label{fig:cem_pipeline}
\vspace{-0.3cm}
\end{figure}
}
  
\newcommand{\figResultsPlan}{
\begin{figure}[!h]
\centering
\includegraphics[width=\linewidth]{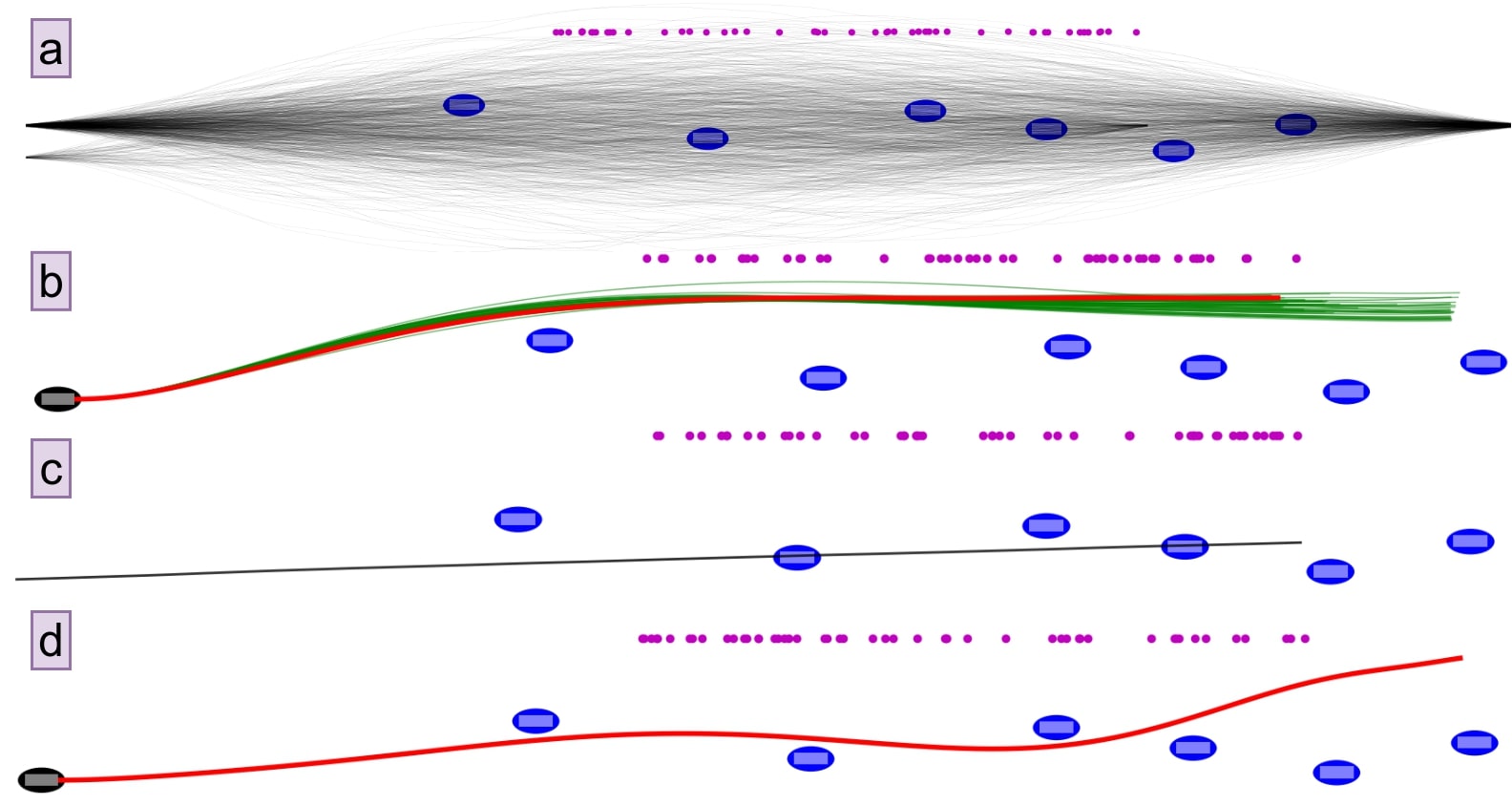}
\setlength{\abovecaptionskip}{-5pt}
\setlength{\belowcaptionskip}{-15pt}
\caption{\small \textbf{Benefit of our CEM-MPC that combines batch non-convex trajectory optimization and adaptive sampling of CEM} \textbf{a}: CEM-MPC trajectory initialization. \textbf{b}: distribution of best performing samples after 10 MPC iterations. \textbf{c}: straight line initialization for a typical trajectory optimizer. \textbf{d}: corresponding optimal trajectory after 10 MPC iterations. Our CEM-MPC is able to converge to a homotopy, bringing the vehicle closer to features (pink points) while avoiding obstacles (blue). In stark contrast, the standard optimizer gets stuck in local minima far from the features.}
\label{fig:resultsPlan}
\end{figure}
}

\newcommand{\figQualResultsControl}{
\begin{figure}[!htb]
\begin{subfigure}[b]{0.23\SubFigureScalar}
    \includegraphics[width=1\linewidth]{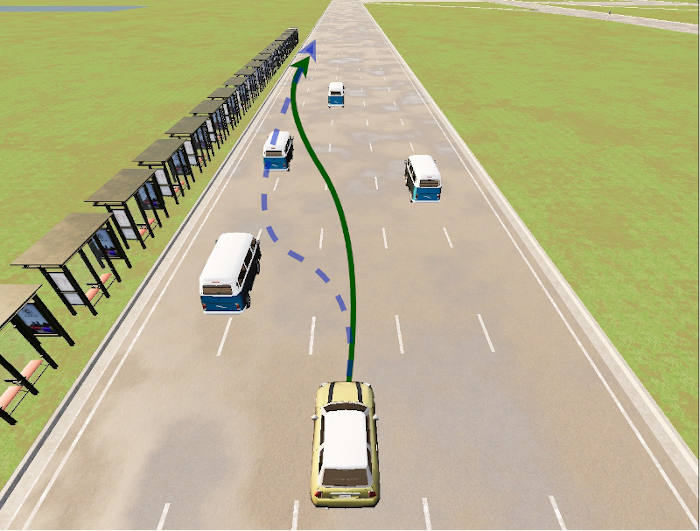}  
\end{subfigure}
\hfill
\begin{subfigure}[b]{0.23\SubFigureScalar}
    \includegraphics[width=1\linewidth]{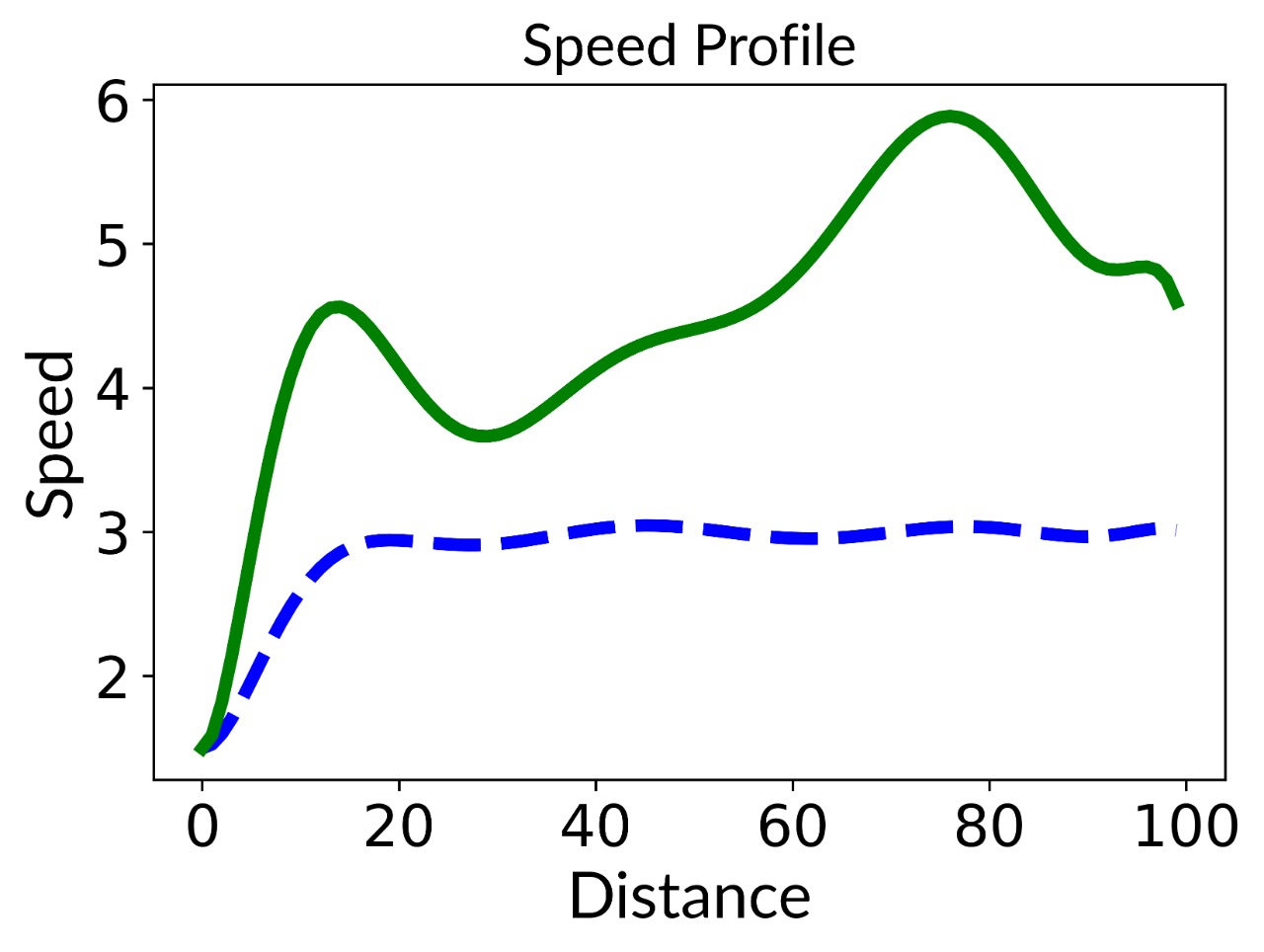}  
\end{subfigure}
\setlength{\belowcaptionskip}{-10pt}
\caption{\captionQualResultsControl}
\label{fig:qualResultsControl}
\end{figure}
}

\newcommand{\figQualResultsGrad}{
\begin{figure}[h]
    \begin{subfigure}[h]{\linewidth}\label{subfig:QualA}
    \includegraphics[width=\linewidth, height=2cm]{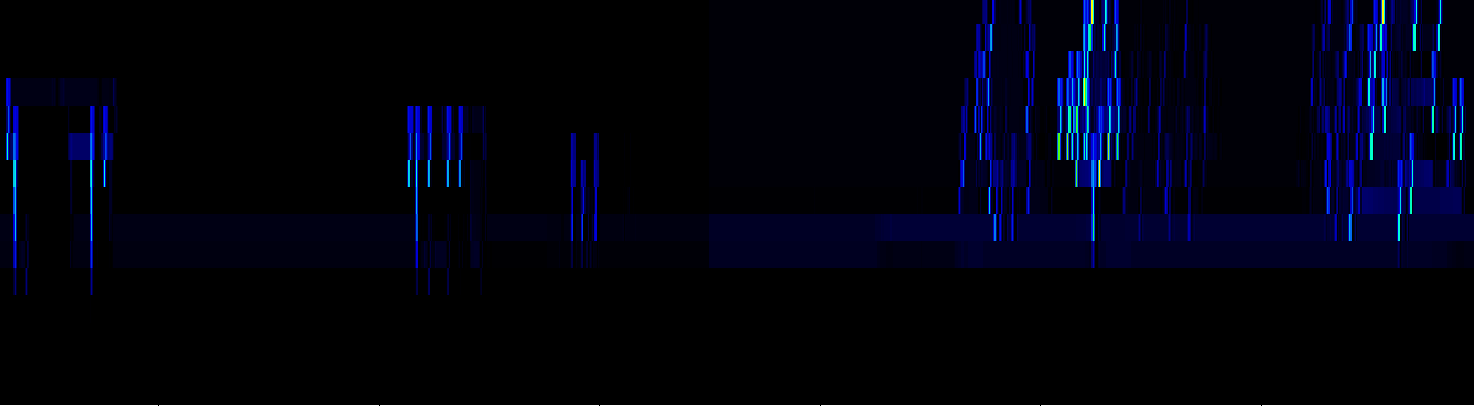}  
\end{subfigure}
\vspace{2px}
\begin{subfigure}[h]{\linewidth}\label{subfig:QualC}
    \includegraphics[width=\linewidth, height=2cm]{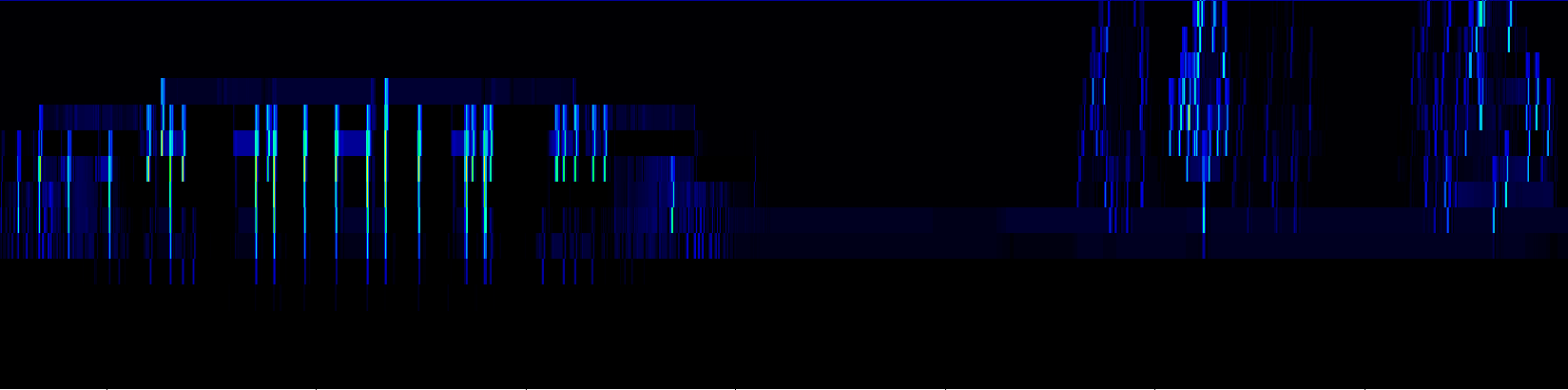}  
\end{subfigure}
\setlength{\abovecaptionskip}{-1pt}
\setlength{\belowcaptionskip}{-5pt}
\caption{\captionQuanResultsGrad}
\label{fig:qualGradCAM}
\end{figure}
}

\newcommand{\figQualResultsAll}{
\begin{figure}[t]
\centering
\includegraphics[width=\linewidth]{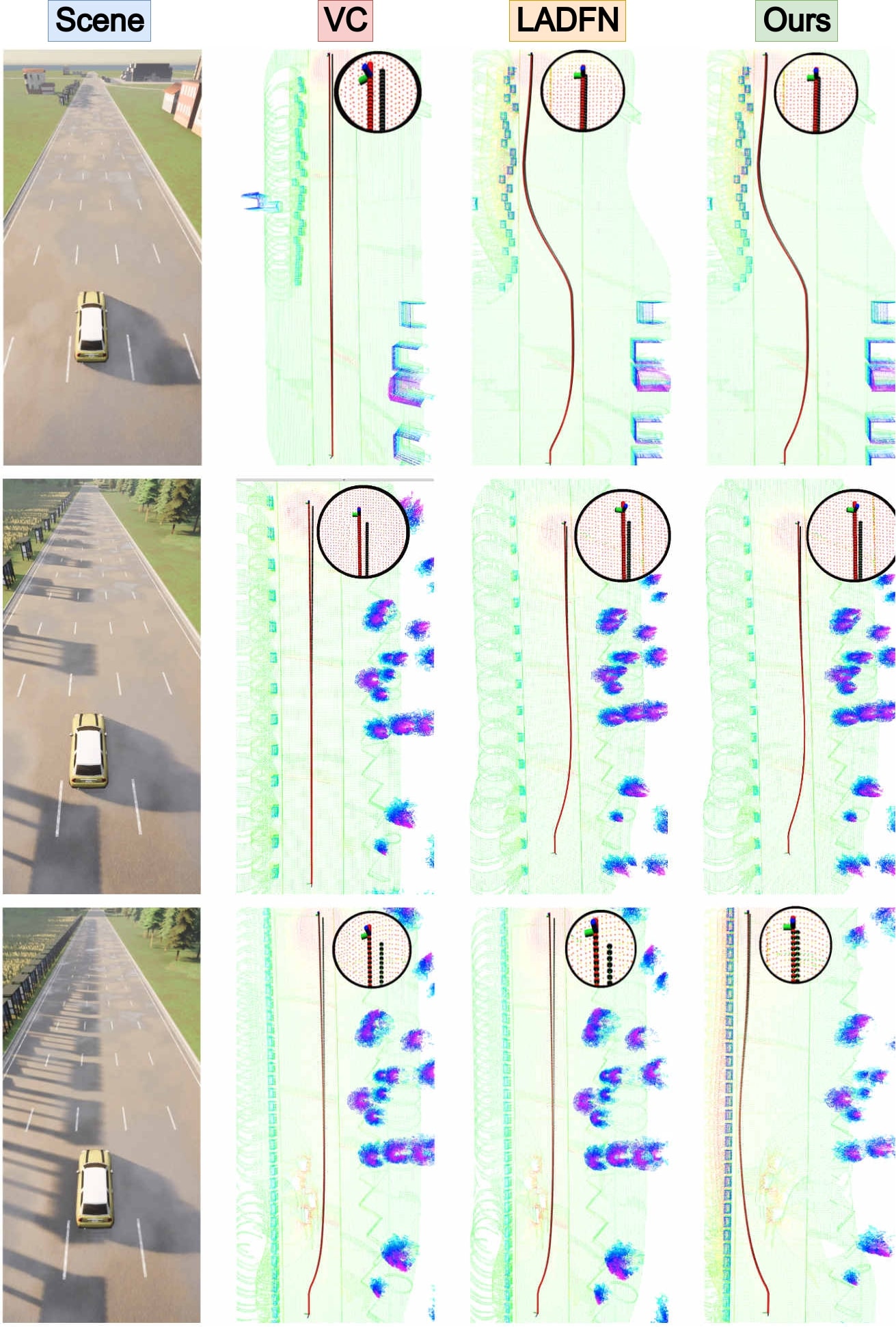}
\setlength{\abovecaptionskip}{-5pt}
\setlength{\belowcaptionskip}{-12pt}
\caption{\captionQualResults}
\label{fig:qualResultsAll}
\vspace*{-2mm}
\end{figure}
}

\newcommand{\figQuanResultsAll}{
\begin{figure}[!htb]
\includegraphics[width=0.5\textwidth]{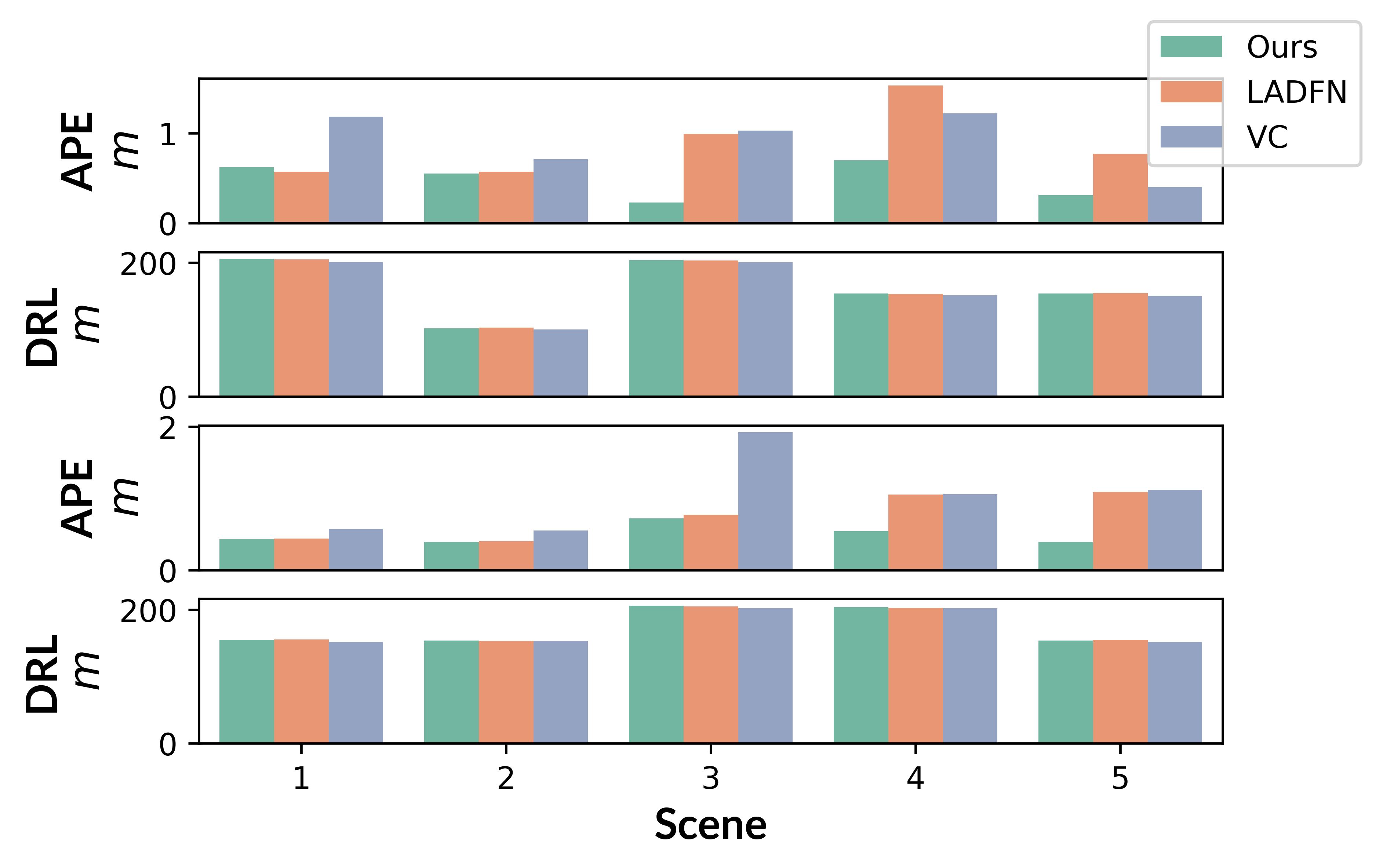}
\centering
\setlength{\abovecaptionskip}{-12pt}
\setlength{\belowcaptionskip}{-15pt}
\caption{\captionQuanResults}
\label{fig:quanResultsAll}
\end{figure}
}

\newcommand{\figAblationResults}{
\begin{figure}[!h]
  \centering
  \setlength{\belowcaptionskip}{-15pt}
  \includegraphics[width=0.8\linewidth]{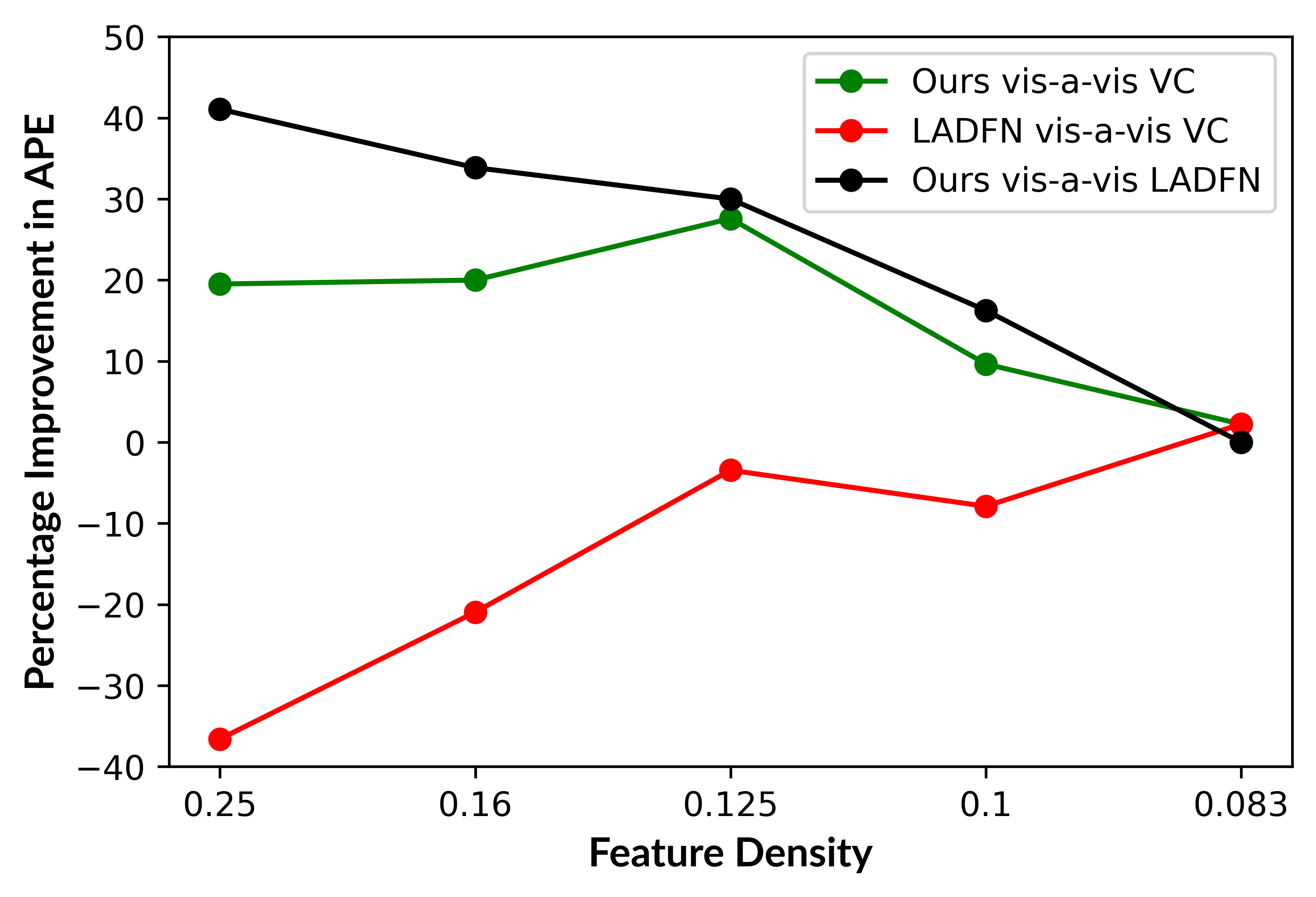}  
  \setlength{\belowcaptionskip}{-15pt}
\caption{\small \textbf{At higher semantic feature densities}, we minimize drift by $\boldsymbol{40\%}$ and $\boldsymbol{20\%}$ over LADFN \cite{omama2021learning} and VC respectively.}
\label{fig:AblationResults}
\end{figure}
}

\setlist[itemize]{leftmargin=*}
\setlist[enumerate]{leftmargin=*}

\title{\LARGE \bf
Drift Reduced Navigation with Deep Explainable Features
}
\author{Mohd Omama$^{1}$, Sundar Sripada V. S.$^{1}$, Sandeep Chinchali$^{2}$, Arun Kumar Singh$^{3}$, K. Madhava Krishna$^{1}$%
\thanks{$^{1}$Robotics Research Center, IIIT-Hyderabad, India}%
\thanks{$^{2}$ECE Department, The University of Texas at Austin}%
\thanks{$^{3}$Institute of Technology, University of Tartu}%
\thanks{The authors thank MathWorks for their generous
financial support}%
}

\begin{document}
\maketitle

\begin{abstract}
Modern autonomous vehicles (AVs) often rely on vision, LIDAR, and even radar-based simultaneous localization and mapping (SLAM) frameworks for precise localization and navigation. However, modern SLAM frameworks often lead to unacceptably high levels of drift (i.e., localization error) when AVs observe few visually distinct features or encounter occlusions due to dynamic obstacles. This paper argues that minimizing drift must be a key desiderata in AV motion planning, which requires an AV to take \textit{active} control decisions to move towards feature-rich regions while also minimizing conventional control cost. To do so, we first introduce a novel data-driven perception module that observes LIDAR point clouds and estimates which features/regions an AV must navigate towards for drift minimization. Then, we introduce an interpretable model predictive controller (MPC) that moves an AV toward such feature-rich regions while avoiding visual occlusions and gracefully trading off drift and control cost. Our experiments on challenging, dynamic scenarios in the state-of-the-art CARLA simulator indicate our method reduces drift up to 76.76\% compared to benchmark approaches. 
\end{abstract}


\section{Introduction}
Precise localization is an important primitive for diverse robots, such as field robots, drones, surgical robots, and especially, self-driving cars. In this paper, we argue that autonomous vehicles should not only optimize for control cost but also a primary drift minimization cost. Often, these are conflicting objectives because the best or shortest trajectory that an ego vehicle can take may not be the most drift-minimizing one (see Fig. \ref{fig:Teaser}). Thus, we require a multi-objective optimization framework to gracefully trade-off both criteria. To the best of our knowledge, no literature so far has embarked on this intersection of work, with the exception of \cite{omama2021learning}. But, \cite{omama2021learning} used an uninterpretable reinforcement learning (RL) approach using Proximal Policy Optimization (PPO) \cite{schulman2017proximal}, which takes many iterations to train, and is difficult to combine with a robust planning framework such as MPC, causing it to generalize poorly to new environments. Hence, this paper takes a principled approach to drastically improve precision robotic navigation. The key observation is that a deep-learned feature module can output drift-minimizing feature locations that can be tracked using sample-efficient, model-based control, thereby closing the gap between interpretable features and a robust planning framework.
\figTeaser

The core novelty in our approach is the use of GradCAM \cite{selvaraju2017grad} to locate drift-minimizing features in the ego vehicle's surroundings. GradCAM \cite{selvaraju2017grad} is an explainable AI tool to determine which input features are most influential for predicting a differentiable output. We also develop a modification of the well-known triplet loss \cite{chechik2010large}, which we call the \textit{Directional Triplet Ranking Loss}, for training our neural network. These improvements to the perception module in our pipeline result in successful identification of features useful for the downstream planning module to minimize drift. 
\figPipeline


Our motion planning module relies on solving tens of hundreds of optimization problems in parallel, each initialized from trajectory samples drawn from a Gaussian distribution. This counters the inherent non-convexity of the problem, escapes poor local minima, and drives the vehicle closer to the features provided by the perception module (see Fig.\ref{fig:resultsPlan}). We combine our batch trajectory optimizer with a Cross-Entropy Method (CEM) style adaptation of the initialization distribution. We run the combined approach in a receding-horizon fashion, which we henceforth call CEM-MPC.


\subsection{Related Work}
Our work broadly relates to active localization in dynamic scenes and perception-aware planning.

\noindent\textit{Active Localization:} Traditional active localization frameworks typically move to regions that are low in state uncertainty. For example, \cite{leung2006active} considers state and observation noise to be Gaussian and selects a trajectory from a candidate set by minimizing the trace of the Covariance Matrix. However, such methods cannot be extended to situations where noise priors are non-Gaussian; formulations in the presence of dynamic actors, such as other cars and pedestrians, become progressively difficult. Initial active localization frameworks in multi-hypothesis settings include \cite{fox1998active}, wherein the agent navigated actively to localize to a unimodal uncertainty from an initial multi-modal hypothesis that typically occurs due to perceptual aliasing. The complexity of active localization was analyzed in \cite{dudek1998localizing} and was shown to be NP-Hard, whereas a randomized algorithm for the same was presented in \cite{rao2005minimum}. The first and possibly only such formulation for active localization in a multi-agent setting was presented in \cite{bhuvanagiri2010motion}. However none of the above frameworks talk of navigating an agent to reduce drift in the presence of moving agents.

\noindent\textit{Perception-Aware Planning:} Recent perception-aware planning methods either aim to keep features with rich texture in a robot's field of view \cite{costante2016perception}, or aim to improve visibility for a maneuver \cite{andersen2019trajectory}, or find appropriate controls to keep a point feature in focus \cite{falanga2018pampc}. These methods also do not account for uncertainty due to dynamic actors nor the semantics of features in their navigation policies.

\noindent\textit{Active Localization in Dynamic Scenes:} The first body of work that solves active localization using LIDAR as an input modality in the presence of dynamic vehicles is \cite{omama2021learning}. Although \cite{omama2021learning} maintains the least drift by moving towards feature-rich regions, it does not address which feature semantics or classes are best for reducing localization error. While the PPO-based RL framework \cite{schulman2017proximal} in \cite{omama2021learning} was successful in reducing drift when compared with typical vehicle controllers such as the Stanley Controller \cite{stanleycontroller}, it does not generalize easily to new scenes nor provide for interpretable controls. 

\subsection{Contributions} \label{sec:contributions}
This work differs from prior literature since we learn drift-minimizing LIDAR features and actively maneuver to them amidst dynamic obstacles, thereby implicitly incorporating semantics and going beyond just feature count. This learning is targeted at the LIDAR Odometry and Mapping (LOAM)\cite{zhang2014loam} SLAM framework which has been the state state of the art for LIDAR odometry and is still, along with its variants, in the top 5 of the KITTI Odometry Benchmark\cite{kitti}. A batch CEM-MPC planner capable of synthesizing hundreds of homotopies in real time integrates the learned features into its selection cost, which chooses the best possible homotopies that reduce drift while simultaneously providing for interpretability. In short, our contributions are as follows: 

\begin{enumerate}
\setlength{\leftmargin}{0pt}
\item \textbf{Identify Features for Drift Minimization using GradCAM:} From the perception input, we have no supervision for identifying and locating features that minimize drift, making it a subtle, indirect problem which is not amenable to supervised learning techniques. Therefore, we infer feature locations by viewing the activation maps of our input range image using GradCAM \cite{selvaraju2017grad}.%
\item \textbf{Directional Triplet Ranking Loss Function:} The subtle nature of the feature identification task necessitates a contrastive loss function that can help train a deep neural network. Hence, we introduce a novel modified Triplet Loss to rank the perception inputs based on how effectively they minimize the ego vehicle's drift.%
\item \textbf{Blending Learning and Classical Control to Beat State-of-the-Art:} Our perception module outputs target features and our interpretable CEM-MPC planner drives the ego-vehicle towards them, reducing drift by up to \textbf{76.76\%} compared to the closest work, LADFN (\underline{L}earning \underline{A}ctions for \underline{D}rift-\underline{F}ree \underline{N}avigation) \cite{omama2021learning}.%
\end{enumerate}
Our LIDAR dataset and the corresponding code can be found at: https://github.com/mohdomama/DRNDEF.

\section{Feature Selection Module} 
The principal goal of our feature selection/perception module is to observe 3D LIDAR point clouds and identify drift-minimizing feature locations. However, the key challenge is that there is no oracle data set or mathematical model that provides ground-truth labels asserting which features are the best at reducing drift. Indeed, this often depends on the complex semantics of a scene and complex nonlinear operations in the LIDAR Odometry and Mapping (LOAM) \cite{zhang2014loam} SLAM framework.

Instead, however, we can obtain a diverse training data set where we start the robot at a certain pose and observe the relative drift at a new pose after it executes a trajectory. Then, our key insight is that we can learn an \textit{embedding} of visual scenes that helps rank whether the drift will increase or decrease if the robot moves to a target pose. Then, we can find which pixels in a visual scene are most responsible for this ranking and move the robot towards these features. We now describe how we (a) learn a representation to rank visual scenes based on drift (\ref{sec:Loss}) and (b) extract the most informative features for this ranking using GradCAM (\ref{sec:GradCAMInference}).

In our system, a VLP-16 LIDAR is attached to an ego vehicle \nameCAR. This collects $N_{\mathrm{pc}}$ 3D point clouds of its surroundings at each discrete time step $t$, namely \nameLIDARINPUT~$\in {\mathbb{R}^{3 \times N_{\mathrm{pc}}}}$. The ego vehicle \nameCAR~incurs a drift \nameDRIFT~at time $t$. We project the 3D point cloud \nameLIDARINPUT~onto a 2D image plane to obtain range image \nameRANGEIMAGE~$\in {\mathbb{R}^{w \times h}}$, where the width and height of the range image $(w,h) = (1800,16)$. From the $N$ raw point clouds \nameLIDARINPUT, we project $N$ such range images \nameRANGEIMAGE~to build our input data set $\mathcal{D}$. This data set $\mathcal{D}$ is used to train a feature extractor to output a rank \nameEMBEDDING~$=$~\nameCNN$(R_t;\theta_{\mathrm{cnn}})$, with the learned parameters $\theta_{\mathrm{cnn}}$. \nameCNN~is a Convolutional Neural Network that takes as input the range image \nameRANGEIMAGE~$\in {\mathbb{R}^{w \times h}}$ and outputs a scalar embedding \nameEMBEDDING~$\in {\mathbb{R}}$. The scalar rank \nameEMBEDDING~represents an ordering of the input range image \nameRANGEIMAGE; higher rank \nameEMBEDDING~means lower drift \nameDRIFT, and vice versa. As previously motivated, identifying drift-minimizing features is an indirect task that we have no supervision over. This requires the need for a ranking-based approach. To obtain this rank \nameEMBEDDING, we train our network \nameCNN~using a novel Directional Triplet Ranking Loss, inspired by \cite{chechik2010large}, and detailed in \ref{sec:TrainingProcedure}. 

Using the input range image \nameRANGEIMAGE~ and the embedding \nameEMBEDDING, we query our network \nameCNN~for obtaining the activation maps \nameGCACTIVATION~$\in {\mathbb{R}^{w \times h}}$ using GradCAM \cite{selvaraju2017grad}. This step is one of the crucial parts of our pipeline as we exploit GradCAM to identify drift-minimizing features percolated through the network. Henceforth, we refer to the output of the Feature Selection module, namely the drift-minimizing features, as simply \textit{features}, as they represent the ``goodness'' of the ego vehicle's perception, and play a key role in the control objective. We now detail the feature selection module.
\vspace{-1.5mm}
\subsection{Range Image Projection} \label{sec:RangeImage}

Range images are a popular proxy representation for LIDAR data. We describe briefly the projection of a LIDAR point cloud input \nameLIDARINPUT~into a range image \nameRANGEIMAGE. Given a point cloud \nameLIDARINPUT~at time $t$ containing the tuple $\langle p_x,p_y,p_z \rangle$ representing each point in Cartesian coordinates, we extract the range (depth) $r$, azimuth angle $\phi$, and elevation angle $\theta$ as: 
\eqnRANGEEXTRACTION
We project this onto an image of width $w$ and height $h$. Using VLP-16's number of channels (16), we set our range image height $h = 16$. Using VLP-16's angular resolution ($0.2^{\circ}$), we set our range image width $w = 1800$ (i.e. $360^\circ$/ $0.2^\circ$). We also require beforehand the vertical field of view of our LIDAR scanner $\mathrm{fov}_{\mathrm{down}} < \mathrm{fov} < \mathrm{fov}_{\mathrm{up}}$. Here, $\mathrm{fov}_{\mathrm{down}}$ and $\mathrm{fov}_{\mathrm{up}}$ are the boundaries of the vertical field of view; for VLP-16, these values are $-15^{\circ}$ and $15^{\circ}$ respectively. The projected image coordinates are:
\vspace{-1.5mm}
\eqnRANGEPROJECTION
where $\mathrm{fov}_{\mathrm{total}} = |\mathrm{fov}_{\mathrm{down}}| + |\mathrm{fov}_{\mathrm{up}}|$. We clamp $x_{\mathrm{proj}}$ and $y_{\mathrm{proj}}$ to the ranges $(0,w-1)$ and $(0,h-1)$ for array indexing. 

\subsection{Directional Triplet Ranking Loss} \label{sec:Loss}
In order to train \nameCNN, we design a new loss called the directional triplet loss that takes inspiration from the standard triplet loss \cite{chechik2010large}.
Here we describe the motivation of designing a new loss function. Contrastive learning facilitates deep neural networks to differentiate between positive and negative samples. With careful selection of samples belonging to positive, negative, and anchor classes $\langle p, n, a \rangle$, we can learn to distinguish between visually similar input images. As previously stated, it is imperative that robots are able to navigate with drift minimization as a primary objective. This criterion informs us that our ego vehicle must change its lateral position intelligently in order to maintain proximity to good features that are conducive to minimizing drift, as established by prior work LADFN \cite{omama2021learning}. Here, lateral refers to the side perpendicular to the road length.

Following this motivation, in practice, we select range images from our data set $\mathcal{D}$~that belong to three different lateral positions on the same road to represent the positive $p$, negative $n$, and anchor $a$~classes of our loss function. We note that these three range images are visually similar and therefore contribute to our motivation for using this loss function. Let us denote the drift incurred by the lateral positions corresponding to the positive, negative, and anchor samples by $d_{p}, d_{n},$ and $d_{a}$, respectively. These drifts satisfy the condition $d_{p} < d_{a} < d_{n}$, i.e., the drift incurred by the positive sample is less than that of the anchor sample, which is less than that of the negative sample. During training, we want our network \nameCNN~to be able to differentiate between the range images \textbf{R}$_p$, \textbf{R}$_n$, and \textbf{R}$_a$ corresponding to the classes in our loss function $\langle p, n, a \rangle$. Hence, we train our network to predict a rank \nameEMBEDDING, which is a scalar in latent space. The ranks corresponding to the positive, negative, and anchor classes are given by $z_p, z_n$ and $z_a$ respectively. We want these ranks to satisfy the condition $z_p > z_a > z_n$, but we also want these values to be close to each other in the latent space since the range images of these three samples are visually similar. 

To ensure that the positive sample has a higher value than the negative sample, we introduce the term $(z_n-z_p)$. To ensure the ranks corresponding to positive and negative samples remain close to the anchor, we use a regularization term $((z_a-z_n)^2 + (z_a-z_p)^2)$. Finally, we maintain the margin $\beta$, present in the standard triplet loss \cite{chechik2010large}, to preclude the rank degeneracy condition $z_p = z_a = z_n$. Therefore, our Directional Triplet Ranking Loss function becomes:
\eqnDTL
\noindent\textit{Significance:} We now describe the benefits of our triplet loss compared to an alternative of predicting absolute drift directly from the ego vehicle's observations. During data collection, it was evident that these absolute drift values had high statistical variation and an extensive range dispersed irregularly throughout our samples. This made it difficult to collect an unbiased data set, skewing the network's results. Further, on any given road, the drift values are similar for different lateral positions, making it difficult to predict the precise value of drift at any road location. 

Our new loss naturally handles these issues. Even with high statistical variance, it is easy to select anchor, positive and negative samples. By choosing these samples from the same road, we can enforce the network to learn the minute differences between range images that are instrumental in improving drift.
Another advantage of our loss is its symphony with GradCAM. As detailed in Sec. \ref{sec:GradCAMInference}, GradCAM can be used to query pixels that maximally activate a differentiable entity, such as the rank \nameEMBEDDING. When we query GradCAM on our network \nameCNN~trained using our directional triplet ranking loss, we obtain the \textit{features} that have the most significant impact in increasing the rank and reducing drift.

\vspace{-1px}
\subsection{Training Procedure} \label{sec:TrainingProcedure}
We collect data set $\mathcal{D}$~using the popular CARLA Simulator \cite{Dosovitskiy17}. An ego vehicle is spawned in various scenes containing features of various semantic classes such as buildings, trees, electric poles, etc. For time interval $t=[0,m]$, where $m$ represents the overall run-time of inference in a single scene $\mathcal{S}$, we run LOAM \cite{zhang2014loam} to perform SLAM and obtain the car's odometry. During this simulation, we record the drift $d_{0:m}$ in the $xy$-plane using the ground truth odometry obtained from CARLA \cite{Dosovitskiy17}, and we also record the point cloud obtained from LOAM, $\mathbf{S}_{0:m}$, to use as the perception input in our pipeline. This is done by spawning the car \nameCAR~initially at time $t=0$, then programatically querying the drifts ${d}_{1}, {d}_{2}, ..., {d}_{m}$ along one lateral position. We repeat this for many lateral positions on the same road to complete the data collection for scene $\mathcal{S}$. This allows us to distinguish the datapoints in one of the sample classes in the triplet loss $\langle p, n, a \rangle$. We collected around 6K total samples, split into 80\% and 20\% for training and validation respectively. 

Our network architecture complements our contrastive loss function. The network is a simple CNN containing 5 blocks; each block contains a convolutional layer, followed by batch normalization and leaky ReLU activation. We perform circular padding on the input tensor's width dimension and zero padding on the height dimension, as this helps range image inputs propagate better through the network. After the 5 blocks, we perform max pooling, flatten the input, and send it through multiple fully connected layers to obtain the rank \nameEMBEDDING. The CNN hyperparameters are as follows: a learning rate of $0.001$, a batch size of $32$, a triplet loss margin $\beta$ of 1, and a circular pad length of $1$ (in the convolutional layers). 
\figCNNIpOp
\subsection{Inferring feature locations using GradCAM} \label{sec:GradCAMInference}
Our ultimate goal is not just to compute a rank, but also the \textit{location} of good features, for which we utilize GradCAM \cite{selvaraju2017grad}. While GradCAM's primary purpose is to  visualize the hidden layers of large CNNs, we exploit it to pinpoint the location of the most influential features on the range image that lead to a high ranking by our network \nameCNN. Specifically, we scope into one of the deeper layers of \nameCNN~ using GradCAM to visualize the activations of our input range image \nameRANGEIMAGE, \nameGCACTIVATION. The activation \nameGCACTIVATION~contains the necessary locations $y_{\mathrm{feat}}$ (re-projected to 3D from \nameGCACTIVATION) needed for our MPC controller to perform trajectory generation and planning. Fig.~\ref{fig:IpOpCNN} shows the input range image \nameRANGEIMAGE~and its corresponding activation \nameGCACTIVATION~after it has percolated through \nameCNN.

\section{Motion Planning Module}
Our planning module is responsible for computing safe, collision-free trajectories that drive the ego-vehicle safely towards the features generated by the perception pipeline described in the previous sections. We follow a trajectory optimization approach under the following assumptions.
\vspace*{-1mm}
\figCEMPipeline

\begin{itemize}
    \item The optimizer operates in the reference-frame attached to the center-line of the road, called the Frenet-Frame \cite{frenet_planner}, which allows us to always plan assuming a straight road. We can map the computed optimal trajectory back to the global frame to align with the road's curvature.
    \item We assume a typical urban driving scenario where the heading of the ego-vehicle with respect to the center-line is small ($\approx \pm 13^{\circ}$) which allows us to bound the shape of the ego-vehicle and obstacles through axis-aligned ellipses without being overly conservative.
\end{itemize}

The mathematical formulation of our trajectory optimizer can be described in the following manner. Here, $(x_{e,t},y_{e, t})$ is the ego-vehicle's position at time $t$. $v_{\mathrm{max}}$ and $a_{\mathrm{max}}$ are the velocity and acceleration bounds, and $y_{\mathrm{feat}}$ are the feature locations supplied by our perception module. $\textbf{b}_0$ and $\textbf{b}_f$ are initial and final boundary conditions. $a$ and $b$ are the major and minor axes of the elliptical approximation of obstacles. $f_v$ and $f_a$ are the speed and acceleration magnitudes.
\eqnObjectiveFunc
\noindent The cost function  \eqref{cost_standard} minimizes (i) the magnitude of accelerations at each time instant, (ii) the lateral separation between the ego-vehicle and the features, and (iii) the ego-vehicles departure from the desired velocity ($v_{\mathrm{des}}$) profile. The equality constraints \eqref{boundary_cond_initial}-\eqref{boundary_cond_final} impose the initial and final boundary conditions. Inequality constraints \eqref{motion_bounds} enforce bounds on the norm of velocity and accelerations while \eqref{coll_nonhol_standar} ensures collision avoidance with the $j^{th}$ obstacle with position $(x_{o, t}^j, y_{o, t}^j)$ at time $t$.

The main complexity of optimization \eqref{cost_standard}-\eqref{coll_nonhol_standar} stems from the non-convex collision avoidance constraints. In the next sub-section, we present our main algorithmic result on the planning side, wherein we combine notions from gradient-free CEM with a batch optimizer that can solve several perturbations of \eqref{cost_standard}-\eqref{coll_nonhol_standar} in parallel.

\subsection{Combining CEM Exploration With Batch Optimization}
\noindent Solving non-convex optimizations like \eqref{cost_standard}-\eqref{coll_nonhol_standar} require an initial guess of the optimal solution. The obtained solution depends on how close this guess is to the actual optimal solution. Our approach leverages this dependency on the initial guess to its advantage. We solve \eqref{cost_standard}-\eqref{coll_nonhol_standar} from different random initializations drawn from a Gaussian distribution and then subsequently adapt the parameters of the distribution based on the resulting optimal trajectory samples. 

Fig.~\ref{fig:cem_pipeline} shows the pipeline of our approach which we refer to as CEM-MPC. It combines the adaptive exploratory properties of CEM with an analytical trajectory optimization to develop a computationally efficient and high performance MPC algorithm. Our CEM-MPC begins (step 1) by initializing a trajectory distribution characterized by a mean $\boldsymbol{\mu}$ and a co-variance $\boldsymbol{\Sigma}$. We draw $n (\approx1000)$ trajectories from this distribution. We then solve $n$ variants of optimization \eqref{cost_standard}-\eqref{coll_nonhol_standar} (step 2) each initialized from a specific trajectory sample drawn in step 1. Since each variant is decoupled from each other, we can solve them in parallel (batch) fashion. The solution process at step 2 results in a distribution of optimal trajectories and we evaluate user-defined meta-costs (to be defined later) on them in step 3. We then rank (step 4) the optimal trajectories based on their associated meta-cost value and choose the top $q$ best performing samples. We then fit a Gaussian distribution to the best performing samples from step 4 to update the $\boldsymbol{\mu}, \boldsymbol{\Sigma}$ for the next MPC iteration. A quintessential problem in CEM is that the variance of the best performing samples could shrink over MPC iterations and this in turn will reduce the diversity in the trajectory initializations. In other words, the exploration ability of the CEM-MPC could deteriorate. We counter this by drawing some trajectory samples from a Gaussian noise and appending them to the best performing samples before executing the mean and covariance update at step 5. The Gaussian noise uses the special co-variance matrix \cite{kalakrishnan2011stomp} which ensures smoothness in the sampled trajectories. The same noise is also used to initialize the MPC algorithm at the first iteration.

\noindent \textbf{Meta-Cost:} Let $(\overline{x}_{e, t}^l, \overline{y}_{e, t}^l), \forall t$ be the $l^{th}$ optimal trajectory obtained from the batch optimization at step 2 of  Fig.~\ref{fig:cem_pipeline}. The meta-cost evaluates the utility of these trajectories:
\eqnMetaCost
\noindent where $f_{\mathrm{curv}}$ represents the curvature function computed based on first and second derivatives of the position trajectories. The term $\kappa_{\mathrm{max}}$ is the maximum permissible curvature and $y_d$ represents the road-width. Thus, our meta-cost \eqref{meta_cost} consists of the primary cost term from the trajectory optimizer \eqref{cost_standard}, and additional penalties that measure the violation on curvature and road boundary constraints.%

\noindent \textbf{Efficient Batch Optimization:} The computational efficiency of our CEM-MPC pipeline rests on how fast we can solve hundreds of non-convex optimizations in parallel (step 2 in Fig.~\ref{fig:cem_pipeline}). To achieve real-time performance, we built a GPU-accelerated variant of our recent batch non-convex optimizer \cite{adajania2022multi} and expanded the mathematical derivation to include the cost functions and constraints specific to the current work.%

\noindent \textbf{An Illustrative Example:} Fig.\ref{fig:resultsPlan} shows a typical result from our CEM-MPC pipeline. \textbf{a} shows the trajectory initialization of our CEM-MPC. \textbf{b} shows the distribution of the top-performing trajectory samples in terms of meta-cost. The best cost trajectory, shown in red, converges to a homotopy that moves the ego-vehicle closer to the pink features. The performance of a standard MPC initialized from a single straight-line trajectory is shown in \textbf{c}. Even though the resulting optimal trajectory is collision-free, it gets stuck in a poor local minimum with a large lateral distance from  features.

\figResultsPlan

\section{Experimental Results} \label{sec:Experiments}

\subsection{Experimental Setup, Metrics and Benchmarks}
Throughout our experimentation, we use the CARLA Simulator \cite{Dosovitskiy17} because its precise physics models of the world and the sensors are ideal for self-driving research. We implemented our CEM-MPC in Python using GPU accelerated linear algebra library JAX \cite{jax2018github}. The typical run-time was around $0.04$s on a i7-8500H laptop with 32GB RAM and RTX2080 Nvidia Graphics card.  

We test our results against two benchmarks. They are the closest work to ours of LADFN (Learning Actions for Drift-Free Navigation) \cite{omama2021learning}, and a single-objective Vanilla (Standard) Stanley Controller (VC). A striking contrast between our approach and LADFN is that the latter uses edge features, calculated geometrically based on a smoothness measure, to identify good features while we have a learned approach for it. We test the three approaches on various scenarios with different run lengths using the following metrics:
\begin{enumerate}[]
    \item \textbf{Absolute Positional Error (APE):} It's a measure of how far off the LOAM pose prediction is with respect to the ground truth.
    \item \textbf{Distance for Run Length (DRL):} For a given run-length along the road, the car can take trajectories of various curvatures, thereby making the distance travelled more than the run length. We call this distance DRL, and it's a measure of the control cost for the three approaches.
\end{enumerate}

We show that we reduce the drift or APE with a slight trade-off in the control cost DRL via drift-aware navigation. Further our system can outperform LADFN in certain scenes where the semantics of features play an essential role.

We perform these tests on 10 scenes that we have created with varying distributions of semantically different features. Five of these scenes don't contain any traffic vehicles and the remaining five contain different configurations of traffic. 

\subsection{Qualitative Results} \label{sec:QualResults}
\figQualResultsAll
Here, we discuss three cases of varying degrees of complexity through which we qualitatively evaluate the effectiveness of our perception system compared to LADFN \cite{omama2021learning} and VC, as shown in Fig.~\ref{fig:qualResultsAll}. Our system performs at par with the previous state-of-the-art (LADFN \cite{omama2021learning}) in simple cases (rows 1 \& 2). In row 3, which is a complex scenario, we outperform both LADFN \cite{omama2021learning} and VC. This is because LADFN \cite{omama2021learning} faces difficulty in assessing the quality of features using its smoothness-based geometric measure, as stated previously. 

\subsubsection{Case 1 - High density of features with no semantic differences}
The scene shown in the first row of Fig.~\ref{fig:qualResultsAll} contains high density of features on one side of the road at a time, and no features on the other side of the road. It is evident that the Vanilla Controller (VC) executes a trajectory with least control cost, whereas LADFN \cite{omama2021learning} and our approach actively move towards the side which contains the higher density of features to minimize drift. 

\subsubsection{Case 2 - Low density of features with semantic differences} Fig.~\ref{fig:qualResultsAll} contains the following: left side contains semantically better features like poles and platforms, but placed far apart, and the right side contains trees that are stacked close together providing LADFN \cite{omama2021learning} with more edge features. In this case, we observe the low density of features on the left side are unable to activate the neurons in our network \nameCNN, while the features on the right side do a better job of neuron activation (Fig.~\ref{fig:qualGradCAM}). Therefore, LADFN and our method move towards the right side. This is further confirmed by our ablation study in \ref{sec:Ablations}. 

\subsubsection{Case 3 - High density of features with semantic differences} This scene shown in the third row of Fig.~\ref{fig:qualResultsAll} is similar to \textit{Case 2}, however the poles and the platforms on the left side are stacked close together and the ego encounters traffic later in the scene. In this case, we observe that the features from the left side maximally activate the neurons of our network \nameCNN~ (Fig.~\ref{fig:qualGradCAM}). As shown in Fig.~\ref{fig:qualResultsAll}, our method moves towards these semantically better features on the left, whereas LADFN still moves the ego car towards the right side that contains more edge features. This results in our approach outperforming both LADFN and VC by a substantial margin of $\boldsymbol{54.2\%}$ and $\boldsymbol{42.6\%}$ respectively.
\figQualResultsGrad

Fig.~\ref{fig:qualResultsControl} shows intelligent behaviours manifested as a result of batch optimization and evaluation of thousands of trajectories in real-time. As shown in green in the left column, the ego vehicle understands to move towards the left, but it also understands that it must first overtake traffic vehicles in front of it; this is reflected aptly in the speed profile on the right (in green). Moreover, the blue trajectory shown on the left assumes that the traffic vehicles are absent. So, the ego vehicle moves towards the feature-rich left side but doesn't expend control cost to perform any overtaking, thereby maintaining a stable speed profile (right, in blue). We note that \cite{omama2021learning} demonstrated similar drift-aware maneuvers using Reinforcement Learning. However, we argue that similar behaviours can be replicated in a more explainable and safer way using analytical generation and evaluation of a large variety of trajectories through a robust MPC method using batch optimization on a GPU.
\figQualResultsControl

\subsection{Quantitative Results} \label{sec:QuanResults}
\figQuanResultsAll
Fig.~\ref{fig:quanResultsAll} shows the trade-off between drift reduction (APE) and control cost (DRL) that our method provides vis-a-vis other benchmarks on static scenes. These experiments are conducted in five diverse scenes with no obstacles, in different towns of the CARLA \cite{Dosovitskiy17} Simulator. In scenes $1$ \& $2$, the trajectories executed by our method and LADFN \cite{omama2021learning} are similar, and hence, the drift improvement that both methods offer over VC are also similar. In scenes $3$ to $5$, our approach results in entirely different, and improved, trajectories compared to LADFN \cite{omama2021learning}. We observe that the features' semantics play a crucial role in drift reduction of up to $\boldsymbol{76.76\%}$ in scenes $3$ to $5$, and our method is able to exploit these semantic differences using GradCAM \cite{selvaraju2017grad}. As previously explained in our qualitative results section \ref{sec:QualResults}, LADFN \cite{omama2021learning} is unable to make use of semantic differences in features, and makes the ego simply move towards regions having higher edge feature density. 

Fig.~\ref{fig:quanResultsAll} shows a similar trade-off between drift reduction (APE) and control cost (DRL) in dynamic scenes. Here, the experiments were conducted in scenes containing various configurations of traffic obstacles. In scenes $1$ to $3$ both LADFN \cite{omama2021learning} and our approach provide similar drift improvement. Scene $3$ contains a low density of features, hence, we note drastic improvement in both LADFN \cite{omama2021learning} and our approach over VC. Once again, when we introduce semantically different features in scenes $4$ and $5$, we observe that our approach outperforms both LADFN \cite{omama2021learning} and VC by up to $\boldsymbol{65.17\%}$ for the aforementioned reasons.

\subsection{Ablation Study on GradCAM Inferences} \label{sec:Ablations}
We conduct an ablation study on feature semantics and their density, shown empirically in Cases 2 \& 3 of \ref{sec:QualResults}. We created a scene containing many edge features on one side, and semantically better features on the other. We have a measure for the density of semantically better features which we vary from $0.25$ (more dense) to $0.083$ (less dense). For density values $0.25$ to $0.125$, we observe that our approach moves the ego vehicle closer to semantically-better features (poles), whereas LADFN \cite{omama2021learning} moves the ego vehicle closer to more edge features (trees). This results in better drift minimization in our approach. As we decrease the density of semantic features, the improvement of our approach over LADFN decreases. At a density value of $.083$ (last row in Table~\ref{tab:AblationResults}), the trajectories of both LADFN and our approach become similar, i.e., both methods moves the ego car closer to the side having more edge features. That is, when the semantically better features are too sparse, they don't provide any substantial improvement (see Fig.~\ref{fig:AblationResults}). Our approach inherently assimilates this understanding.
\tableAblationResults

\figAblationResults
 
\section{Conclusion}
This paper presents a method for autonomous vehicles to significantly reduce drift while navigating in complex, dynamic scenes by optimizing jointly for feature proximity and control cost. Our key innovation is to learn an embedding of scenes through a novel contrastive loss function that helps distinguish features that are beneficial for drift minimization. Our experiments illustrate how our method tightly integrates learned perception with interpretable MPC to reduce drift up to $\boldsymbol{76.76\%}$ in challenging driving scenarios. 

In future work, we plan to deploy our drift minimizing MPC controller on a research prototype autonomous vehicle. Moreover, we plan to deploy the same algorithm on a small indoor robot equipped with a Velodyne LIDAR sensor in crowded indoor scenes with visual occlusions and dynamic agents. Finally, we plan to extend the same core algorithm to navigate robots in a factory floor while always maintaining sufficient proximity and connectivity to WiFi access points, which is important in networked/cloud robotics applications. 

\bibliographystyle{IEEEtran}
\bibliography{bibs}

\begin{thebibliography}{10}
\providecommand{\url}[1]{#1}
\csname url@samestyle\endcsname
\providecommand{\newblock}{\relax}
\providecommand{\bibinfo}[2]{#2}
\providecommand{\BIBentrySTDinterwordspacing}{\spaceskip=0pt\relax}
\providecommand{\BIBentryALTinterwordstretchfactor}{4}
\providecommand{\BIBentryALTinterwordspacing}{\spaceskip=\fontdimen2\font plus
\BIBentryALTinterwordstretchfactor\fontdimen3\font minus
  \fontdimen4\font\relax}
\providecommand{\BIBforeignlanguage}[2]{{%
\expandafter\ifx\csname l@#1\endcsname\relax
\typeout{** WARNING: IEEEtran.bst: No hyphenation pattern has been}%
\typeout{** loaded for the language `#1'. Using the pattern for}%
\typeout{** the default language instead.}%
\else
\language=\csname l@#1\endcsname
\fi
#2}}
\providecommand{\BIBdecl}{\relax}
\BIBdecl

\bibitem{omama2021learning}
M.~Omama, S.~S. Venugopalaswamy~Sriraman, S.~Chinchali, and K.~Madhava~Krishna,
  ``Ladfn: Learning actions for drift-free navigation in highly dynamic
  scenes,'' (in press) American Control Conference 2022.

\bibitem{schulman2017proximal}
J.~Schulman, F.~Wolski, P.~Dhariwal, A.~Radford, and O.~Klimov, ``Proximal
  policy optimization algorithms,'' \emph{arXiv preprint arXiv:1707.06347},
  2017.

\bibitem{Dosovitskiy17}
A.~Dosovitskiy, G.~Ros, F.~Codevilla, A.~Lopez, and V.~Koltun, ``{CARLA}: {An}
  open urban driving simulator,'' in \emph{Proceedings of the 1st Annual
  Conference on Robot Learning}, 2017, pp. 1--16.

\bibitem{selvaraju2017grad}
R.~R. Selvaraju, M.~Cogswell, A.~Das, R.~Vedantam, D.~Parikh, and D.~Batra,
  ``Grad-cam: Visual explanations from deep networks via gradient-based
  localization,'' in \emph{Proceedings of the IEEE international conference on
  computer vision}, 2017, pp. 618--626.

\bibitem{chechik2010large}
G.~Chechik, V.~Sharma, U.~Shalit, and S.~Bengio, ``Large scale online learning
  of image similarity through ranking.'' \emph{Journal of Machine Learning
  Research}, vol.~11, no.~3, 2010.

\bibitem{leung2006active}
C.~Leung, S.~Huang, and G.~Dissanayake, ``Active slam using model predictive
  control and attractor based exploration,'' in \emph{2006 IEEE/RSJ
  International Conference on Intelligent Robots and Systems}.\hskip 1em plus
  0.5em minus 0.4em\relax IEEE, 2006, pp. 5026--5031.

\bibitem{fox1998active}
D.~Fox, W.~Burgard, and S.~Thrun, ``Active markov localization for mobile
  robots,'' \emph{Robotics and Autonomous Systems}, vol.~25, no. 3-4, pp.
  195--207, 1998.

\bibitem{dudek1998localizing}
G.~Dudek, K.~Romanik, and S.~Whitesides, ``Localizing a robot with minimum
  travel,'' \emph{SIAM Journal on Computing}, vol.~27, no.~2, pp. 583--604,
  1998.

\bibitem{rao2005minimum}
M.~Rao, G.~Dudek, and S.~Whitesides, ``Minimum distance localization for a
  robot with limited visibility,'' in \emph{Proceedings of the 2005 IEEE
  International Conference on Robotics and Automation}.\hskip 1em plus 0.5em
  minus 0.4em\relax IEEE, 2005, pp. 2438--2445.

\bibitem{bhuvanagiri2010motion}
S.~Bhuvanagiri and K.~M. Krishna, ``Motion in ambiguity: Coordinated active
  global localization for multiple robots,'' \emph{Robotics and Autonomous
  Systems}, vol.~58, no.~4, pp. 399--424, 2010.

\bibitem{costante2016perception}
G.~Costante, C.~Forster, J.~Delmerico, P.~Valigi, and D.~Scaramuzza,
  ``Perception-aware path planning,'' \emph{arXiv preprint arXiv:1605.04151},
  2016.

\bibitem{andersen2019trajectory}
H.~Andersen, J.~Alonso-Mora, Y.~H. Eng, D.~Rus, and M.~H. Ang, ``Trajectory
  optimization and situational analysis framework for autonomous overtaking
  with visibility maximization,'' \emph{IEEE Transactions on Intelligent
  Vehicles}, vol.~5, no.~1, pp. 7--20, 2019.

\bibitem{falanga2018pampc}
D.~Falanga, P.~Foehn, P.~Lu, and D.~Scaramuzza, ``Pampc: Perception-aware model
  predictive control for quadrotors,'' in \emph{2018 IEEE/RSJ International
  Conference on Intelligent Robots and Systems (IROS)}.\hskip 1em plus 0.5em
  minus 0.4em\relax IEEE, 2018, pp. 1--8.

\bibitem{stanleycontroller}
G.~M. Hoffmann, C.~J. Tomlin, M.~Montemerlo, and S.~Thrun, ``Autonomous
  automobile trajectory tracking for off-road driving: Controller design,
  experimental validation and racing,'' in \emph{2007 American Control
  Conference}, 2007, pp. 2296--2301.

\bibitem{zhang2014loam}
J.~Zhang and S.~Singh, ``Loam: Lidar odometry and mapping in real-time.'' in
  \emph{Robotics: Science and Systems}, vol.~2, no.~9, 2014.

\bibitem{kitti}
A.~Geiger, P.~Lenz, and R.~Urtasun, ``Are we ready for autonomous driving? the
  kitti vision benchmark suite,'' in \emph{Conference on Computer Vision and
  Pattern Recognition (CVPR)}, 2012.

\bibitem{frenet_planner}
M.~Werling, J.~Ziegler, S.~Kammel, and S.~Thrun, ``Optimal trajectory
  generation for dynamic street scenarios in a frenet frame,'' in \emph{2010
  IEEE International Conference on Robotics and Automation}.\hskip 1em plus
  0.5em minus 0.4em\relax IEEE, 2010, pp. 987--993.

\bibitem{kalakrishnan2011stomp}
M.~Kalakrishnan, S.~Chitta, E.~Theodorou, P.~Pastor, and S.~Schaal, ``Stomp:
  Stochastic trajectory optimization for motion planning,'' in \emph{2011 IEEE
  international conference on robotics and automation}.\hskip 1em plus 0.5em
  minus 0.4em\relax IEEE, 2011, pp. 4569--4574.

\bibitem{adajania2022multi}
V.~K. Adajania, A.~Sharma, A.~Gupta, H.~Masnavi, M.~Krishna, and A.~K. Singh,
  ``Multi-modal model predictive control through batch non-holonomic trajectory
  optimization: Application to highway driving,'' \emph{IEEE Robotics and
  Automation Letters}, 2022.

\bibitem{jax2018github}
\BIBentryALTinterwordspacing
J.~Bradbury, R.~Frostig, P.~Hawkins, M.~J. Johnson, C.~Leary, D.~Maclaurin,
  G.~Necula, A.~Paszke, J.~Vander{P}las, S.~Wanderman-{M}ilne, and Q.~Zhang,
  ``{JAX}: composable transformations of {P}ython+{N}um{P}y programs,'' 2018.
  [Online]. Available: \url{http://github.com/google/jax}
\BIBentrySTDinterwordspacing

\end{thebibliography}
\end{document}